\documentclass{article}
\usepackage[T1]{fontenc}
\usepackage[numbers]{natbib}  
\usepackage{numprint}  
\usepackage{graphicx} 
\usepackage{xcolor} 
\usepackage{tabularx}  
\usepackage{bbm, amsmath, amssymb}
\usepackage{authblk}
\usepackage[acronyms]{glossaries}
\usepackage{booktabs}
\usepackage{hyperref}

\usepackage{url}

\newacronym{AL}{AL}{Active Learning}
\newacronym{ANN}{ANN}{Artificial Neural Network}
\newacronym{AOI}{AOI}{area of interest}
\newacronym{API}{API}{Application Programming Interface}
\newacronym{BC}{BC}{British Columbia}
\newacronym{BERT}{BERT}{Bidirectional Encoder Representations from Transformers}
\newacronym{Bi-LSTM}{Bi-LSTM}{Bidirectional LSTM}
\newacronym{BT}{BT}{Breaking Ties}
\newacronym{CAMM}{CAMM}{Cross-Attention Multi-Modal}
\newacronym{CEMS}{CEMS}{Copernicus Emergency Management Service}
\newacronym{CNN}{CNN}{Convolutional Neural Network}
\newacronym{DAL}{DAL}{Discriminative Active Learning}
\newacronym{EO}{EO}{Earth Observation}
\newacronym{ERCC}{ERCC}{Emergency Response Coordination Centre}
\newacronym{GCS}{GCS}{Greedy Core-Set}
\newacronym{GDACS}{GDACS}{Global Disaster Alert and Coordination System}
\newacronym{GI}{GI}{Geographic Information}
\newacronym{GFT}{GFT}{Generic Fine-Tuning}
\newacronym{GMLZ}{GMLZ}{German Joint Information and Situation Centre}
\newacronym{GNN}{GNN}{Graph Neural Network}
\newacronym{KWF}{KWF}{Keyword-filtering}
\newacronym{HAN}{HAN}{Hierarchical Attention Network}
\newacronym{i.i.d.}{i.i.d.}{independent and identically distributed}
\newacronym{LC}{LC}{Least Confidence}
\newacronym{LLM}{LLM}{Large Language Model}
\newacronym{LSTM}{LSTM}{Long Short-Term Memory}
\newacronym{KNN}{KNN}{k-nearest neighbour}
\newacronym{LDA}{LDA}{Latent Dirichlet Allocation}
\newacronym{ML}{ML}{Machine Learning}
\newacronym{NB}{NB}{Naive Bayes}
\newacronym{PE}{PE}{Prediction Entropy}
\newacronym{RF}{RF}{Random Forest}
\newacronym{RMS}{RMS}{Rapid Mapping Service}
\newacronym{RoBERTa}{RoBERTa}{Robustly Optimized BERT Pre-training Approach}
\newacronym{SEM}{SEM}{Satellite-based emergency mapping}
\newacronym{SVM}{SVM}{Support Vector Machine}
\newacronym{SENAPRED}{SENAPRED}{National System for Disaster Prevention, Mitigation and Attention}
\newacronym{TF-IDF}{TF-IDF}{Term Frequency-Inverse Document Frequency}
\newacronym{UCPM}{UCPM}{EU Civil Protection Mechanism}
\newacronym{WWW}{WWW}{World Wide Web}

\title{Active Learning for Identifying Disaster-Related Tweets: A Comparison with Keyword Filtering and Generic Fine-Tuning}
\author[1]{David Hanny}
\author[1]{Sebastian Schmidt}
\author[1,2]{Bernd Resch}
\affil[1]{Department of Geoinformatics - Z\_GIS, University of Salzburg, Austria}
\affil[2]{Center for Geographic Analysis, Harvard University, US}
\date{}                     
\begin{document}
\maketitle              

\begin{abstract}
Information from social media can provide essential information for emergency response during natural disasters in near real-time. However, it is a difficult task to identify the disaster-related posts among the large amounts of unstructured data available. Previous methods often use keyword filtering, topic modelling or classification-based techniques to identify such posts. \acrfull{AL} presents a promising sub-field of \acrfull{ML} that has not been used much in the field of text classification of social media content. This study therefore investigates the potential of \acrshort{AL} for identifying disaster-related Tweets. We compare a keyword filtering approach, a RoBERTa model fine-tuned with generic data from CrisisLex, a base RoBERTa model trained with \acrshort{AL} and a fine-tuned RoBERTa model trained with \acrshort{AL} regarding classification performance. For testing, data from CrisisLex and manually labelled data from the 2021 flood in Germany and the 2023 Chile forest fires were considered. The results show that generic fine-tuning combined with 10 rounds of \acrshort{AL} outperformed all other approaches. Consequently, a broadly applicable model for the identification of disaster-related Tweets could be trained with very little labelling effort. The model can be applied to use cases beyond this study and provides a useful tool for further research in social media analysis.

\end{abstract}

\section{Introduction}
\label{sec:introduction}
Identifying important contents among vast amounts of information has been a central research topic in many disciplines including the digital and analytical sciences. Berners-Lee \cite{Berners-Lee.1997} argues that "[i]nformation about information is powerful not just as information, but because it allows one to leverage one's use of information". As a result, numerous methods have been developed to query and filter various types of data as efficiently as possible \cite{Chowdhury.2010, Hanani.2001}.

Simultaneously, social media has become one of the most prevalent and abundant data sources in modern times \cite{Newman.2023}. Geo-referenced social media data in particular has proven to be a vital source of data before, during and after the occurrence of natural disasters. Emergency responders, official entities and aid organisations can use social media platforms to intercept information in near real-time. This includes texts and imagery with potentially valuable information \cite{Luna.2018}. Given the complexity of natural language, even seemingly simple tasks such as identifying the posts that are related to the disaster pose a challenging task. To solve it, numerous methods have already been proposed in the literature, ranging from naive keyword filtering \cite{Sarker.2020} to advanced techniques such as topic modelling \cite{Havas.2021} or classification using neural networks \cite{Madichetty.2019}. Overall, the majority of methods in the current literature can be categorised as \gls{ML}.

Since the identification of disaster-related Tweets is generally a binary classification task, the use of unsupervised techniques such as topic modelling might not yield consistent results because they are highly dependent on the data set \cite{Agrawal.2018}. Simultaneously, many supervised learning approaches require considerable amounts of training data, resulting in extensive labelling efforts to train a high-quality model. For this reason, \acrfull{AL} has gained increasing popularity. It describes a semi-supervised learning approach in which the model chooses the samples to label instead of humans. This minimises the amount of labelled data needed to achieve high model performance \cite{Lewis.1994}. Throughout \gls{AL}, borderline cases are also presented to the annotator which can guide and thereby improve the learning process.

So far, \gls{AL} has rarely been used in the context of social media analysis. Paul et al. \cite{Paul.2023} show that it can yield promising results for the identification of disaster-related Tweets. However, their study is limited to one data set and they do not explicitly evaluate their outputs for different kinds of natural disasters. Furthermore, their \gls{AL} approach is not compared to other methods of filtering. To fill this research gap, we aim to answer the following research question: How does an \gls{AL}-based approach compare to keyword filtering or fine-tuning using a broad generic data set for the classification of disaster-related Tweets?

\section{Related work}
\label{sec:related_work}
The previous work related to this study concerns the classification of disaster-related social media posts and \gls{AL} for textual data.

\subsection{Classification of disaster-related social media posts}
Data from various social media platforms has proven useful in the context of disaster management. Numerous methods have already been developed for the detection of events, e.g. by Saeed et al. \cite{Saeed.2019}.
Some of these approaches rely on a simple keyword-filtering in combination with thresholds \cite{Shah.2021} or are based on disaster-related hashtags \cite{RayChowdhury.2020}. Chen et al. \cite{Chen.2018} introduce a keyword-based query strategy that iteratively updates the keyword list by ranking \gls{TF-IDF} weights from previously identified posts. Yigitcanlar et al. \cite{Yigitcanlar.2022} base their analysis of disaster-related Tweets in Australia on word frequency and co-occurrence.

More advanced approaches based on \gls{ML} algorithms have also been devised.
Havas et al. \cite{Havas.2021} use the probabilistic \gls{LDA} method to model topics in Tweets for near real-time monitoring of natural disasters. This approach is also used by Sit et al. \cite{Sit.2019} in combination with a \gls{LSTM} network to identify topics in Tweets about Hurricane Irma. Saleem et al. \cite{Saleem.2023} use BERTopic, a topic modelling approach based on \gls{BERT}, to identify relevant topics for the 2023~Turkey earthquake.
Models from the \gls{BERT} family have generally been employed frequently in this domain. For example, de Bruijn et al. \cite{deBruijn.2019} utilise \gls{BERT} to detect flood events from social media posts.
Huang et al. \cite{Huang.2022} identify emergency-related posts on Sina Weibo by putting semantic representations from \gls{BERT} into a \gls{Bi-LSTM} network with an attention mechanism. 
One frequently used variant of \gls{BERT} is the \gls{RoBERTa} which is based on an improved pre-training procedure resulting in an even better understanding of natural language \cite{Liu.2019}. A \gls{RoBERTa} model fine-tuned with the CrisisMMD data set is used by Koshy et al. \cite{Koshy.2023} in combination with a Vision Transformer model for imagery to categorise multimodal Twitter data. 
Madichetty et al. \cite{Madichetty.2023} use \gls{RoBERTa} and VGG-16 to classify textual and imagery content for various disasters including hurricanes and wildfires. Multiplying the output class probabilities, they fuse this information to identify informative Tweets. Adwaith et al. \cite{Adwaith.2022} compare multimodal architectures for text and imagery, identifying a combination of \gls{RoBERTa} and ResNet or DenseNet as most suitable.
Additionally, a \gls{CNN} has been proposed by several authors, e.g. for anomaly detection in a global Twitter feed \cite{Sufi.2022} or to identify landslide imagery from social media \cite{Pennington.2022}.
\gls{GNN}-based methods have also been developed, e.g. by Li et al. \cite{Li.2022b} who combine textual and imagery content for their classifications. Papadimos et al. \cite{Papadimos.2023} additionally include the timestamp into their \gls{GNN}.
Some of these methods have also been employed for the more specific task of relevance classification. Madichetty et al. \cite{Madichetty.2019}, for instance, develop a \gls{CNN} to categorise disaster-related Tweets as either informative or uninformative. 

\subsection{Active learning for textual data}
\gls{AL} is a popular method for semi-supervised classification, which aims to minimise the amount of labelled data for training by allowing the machine to choose suitable training examples from a larger collection of unlabelled data. Those samples are then annotated by an oracle, generally involving a human-in-the-loop. Subsequently, the model is updated using the newly labelled data \cite{Lewis.1994, Mosqueira-Rey.2023}. With the help of \gls{AL}, a relatively high model accuracy can be achieved with little training data as redundant data points can be avoided by the appropriate query strategy. This, in turn, drives down the cost of labelling data \cite{Settles.2009}. 

A variety of query strategies have been developed for \gls{AL}. Monarch and Manning \cite{Monarch.2021} distinguishes three types of strategies: random, uncertainty (selecting the instances with the lowest label certainty) and diversity (selecting the instances that are rare in the training data to broaden the training space). Implementations of those strategies that have successfully been utilised for language classification include \cite{Ein-Dor.2020, Schroder.2023a}:

\begin{itemize}
    \item \textbf{Random sampling}: Batch instances are chosen at random from the unlabelled training data set.
    \item \textbf{\gls{LC}}: Selects the instances with the least prediction confidence \cite{Lewis.1994}.
    \item \textbf{\gls{PE}}: Chooses instances that maximise the expected amount of information we gain about the set of unlabelled data points \cite{Holub.2008}.
    \item \textbf{\gls{BT}}: Selects the instances that have the smallest margin between their most likely and second most likely predicted class, i.e. the ties \cite{Luo.2005}.
    \item \textbf{\gls{GCS}}: Greedily selects the instances that best cover the data set in the learned representation space using the geometry of the data points \cite{Sener.2018}.
    \item \textbf{\gls{DAL}}: Selects instances that make the batch most representative of the entire pool of unlabelled training instances using a binary classification setting \cite{Gissin.2019}.
\end{itemize}

Ein-Dor et al. \cite{Ein-Dor.2020} examine different query strategies for \gls{AL} with \gls{BERT}. While all \gls{AL} strategies improve over the baseline of choosing samples randomly, no single strategy consistently outperforms all its counterparts. Nonetheless, various strategies yield different runtimes, sample diversity and representativeness concerning the full unlabelled pool of training data. \gls{DAL} and \gls{GCS} have been shown to deliver the most diverse batches. \gls{DAL} additionally yields higher representativeness when compared to \gls{GCS}. However, no direct connection between these measures and classification performance has been demonstrated, giving these results only theoretical value. Lowell et al. \cite{Lowell.2019} reveal the inconsistency of \gls{AL} over \gls{i.i.d.} sampling more explicitly. In a greater research context, \gls{AL} has successfully been utilised and examined for a diverse range of tasks in natural language and image processing. Sahan et al. \cite{Sahan.2021} compare different types of uncertainty representations for text classification and fake news detection. Budd et al. \cite{Budd.2021} investigate the role of humans in the development and integration of deep learning models for medical image analysis using \gls{AL}. Ahmed et al. \cite{Ahmed.2020} show that \gls{AL} can be equally useful in centralised and federated learning environments. They further demonstrate that \gls{AL} can be utilised independently of the data set by evaluating their method for natural disaster image and waste classification. For both applications, their \gls{AL} method achieves competitive results when compared to the scenario that each sample is manually analysed and annotated.

\section{Methodology}
\label{sec:methodology}
\begin{figure}[ht]
    \centering
    \includegraphics[width=\textwidth]{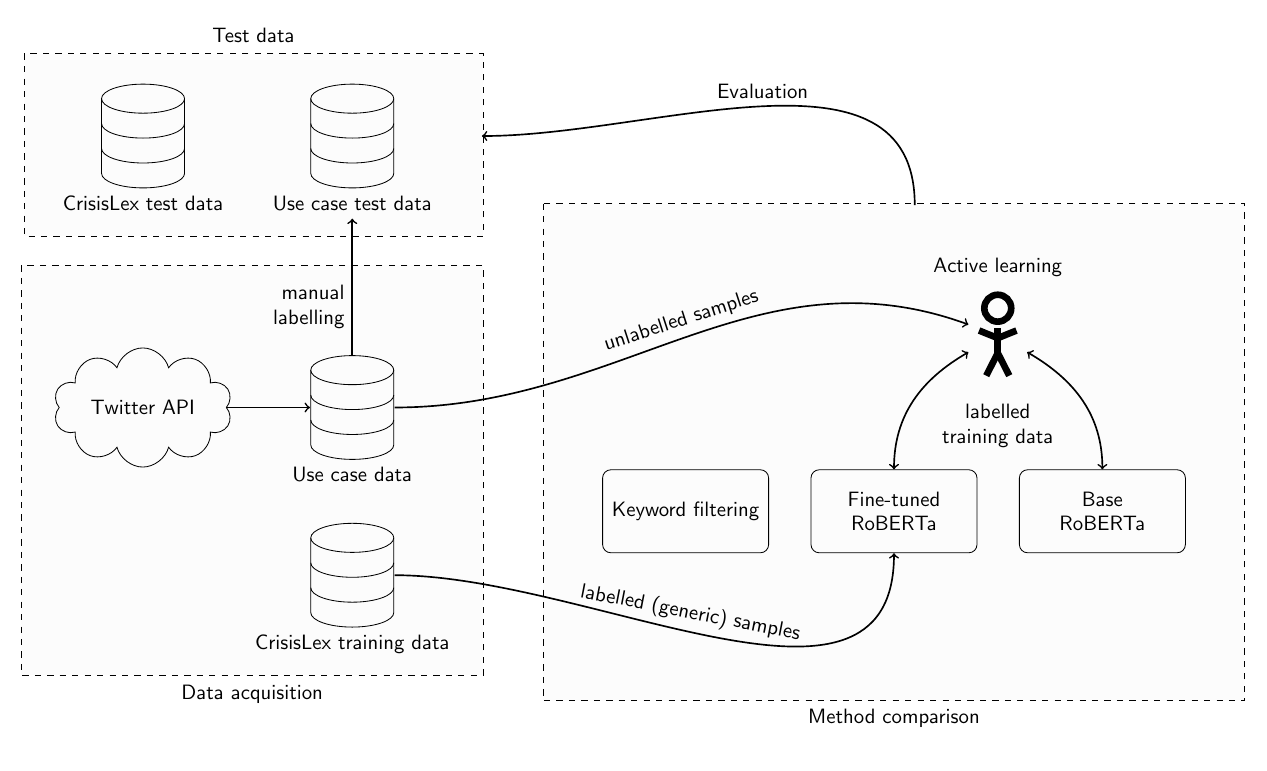}
    \caption{Workflow of the study including data collection and methodology}
    \label{fig:workflow}
\end{figure}

In the following section, we provide an overview of our data and methodological approach. Fig.~\ref{fig:workflow} shows an outline of our workflow. 

\subsection{Data}
\label{sec:data}

The social media platform Twitter (now officially X) provides data through various \gls{API} endpoints. We followed Havas et al. \cite{Havas.2021b} and retrieved georeferenced Tweets using both the v1.1 REST and the streaming \gls{API}.
To evaluate the performance of Tweet classification with \gls{AL}, we decided to go with two use cases: In July~2021, one of the most devastating floods in recent German history took place in the Ahr valley in Rhineland-Palatinate \cite{Fekete.2021}. In early~2023, extensive forest fires raged in the Central South of Chile, particularly in the Ñuble, Biobío and La~Araucanía regions.
To create our training and test data sets, we employed spatial and temporal filtering (cf. Table~\ref{tab:training_data}).
Since the \gls{AOI} and timeframe were much larger for Chile, we obtained substantially more Tweets. For further processing, the data set was therefore pre-filtered using the keywords "incendio", "fuego" and "fire" to narrow down the search space for disaster-related Tweets. For evaluation, a fraction of the two data sets were manually labelled as "related" or "unrelated" to the disaster by two annotators with uniform inter-annotator agreement. In this study, a Tweet was only considered disaster-related if it contained reactions, impressions, commentaries or other explicit information about the respective natural disaster.

\begin{center}
\begin{table}[ht]
\caption{Properties of our use case data sets}
\begin{tabular}{p{3.35cm}p{3.75cm}p{2cm}p{2.5cm}}
\toprule
\textbf{Use case} & \textbf{Timeframe} &  \textbf{\#Tweets} & \textbf{\#Test data} \\
\midrule
2021 Germany flood      &  2021-07-01 - 2021-08-01 & \numprint{11175} & 192 \\  
2023 Chile forest fires & 2023-01-01 - 2023-04-30 &  \numprint{1739986} & 364  \\ 
\bottomrule
\end{tabular}
\label{tab:training_data}
\end{table}
\centering
\end{center}

Moreover, a generic data set of disaster-related and non-disaster-related Tweets was compiled that did not require any manual annotation. It was built upon data sets from CrisisLexT6 \cite{Olteanu.2014} and CrisisLexT26 \cite{Olteanu.2015}, for which a label mapping as in Table~\ref{tab:label_mappings} was created. Based on the re-labelled data sets, a class-balanced training data set was curated. Non-natural disasters (e.g. shootings or bombings) were excluded from the training data as they were irrelevant to the use cases in question. The final data set contained Tweets regarding earthquakes, volcanic eruptions, landslides, wildfires, floods and tropical storms. To further augment the data sets and improve the multilingual capabilities during training, the curated English-language training data was translated to Spanish, German, Italian and French using the M2M-100 translation model \cite{Fan.2020} to obtain a multilingual joint data set consisting of \numprint{179391} training data points and \numprint{44848} test data points. 

\begin{center}
\begin{table}[ht]
\caption{Label mappings used to curate the training data}
    \centering
    \begin{tabular}{llll}
        \toprule
        \multicolumn{2}{c}{\textbf{CrisisLexT6}} & \multicolumn{2}{c}{\textbf{CrisisLexT26}} \\
         Label & New label & Informativeness label & New label \\
         \midrule
         on-topic & related & Related and informative & related \\
         & & Related - but not informative & related\\
         off-topic & unrelated & Not related & unrelated \\
         & & Not applicable & - \\
         \bottomrule
    \end{tabular}
    \label{tab:label_mappings}
\end{table}
\centering
\end{center}

\subsection{Keyword filtering}
Keyword filtering has been utilised in a number of studies to identify Tweets related to a specific topic within a much larger collection \cite{Herfort.2014, Sarker.2020, Shah.2021}. Due to its simplicity and transparency, it was also evaluated as a filtering technique for this study. It was implemented using a basic containment check for each keyword that was applied to each string while casing was ignored. Table~\ref{tab:keywords} shows the keywords used for each test data set. As the data sets contain Tweets in multiple languages, the keywords were translated to all of the listed languages.

\begin{center}
\begin{table}[ht]
\caption{List of keywords used for keyword-filtering for the respective test data set}
\centering
\begin{tabularx}{\textwidth}{Xll}
\toprule
\textbf{Keywords} & \textbf{Languages} & \textbf{Data set} \\
\midrule
earthquake, volcano, landslide, fire, flood, tornado, typhoon, erdbeben, vulkan, erdrutsch, feuer, flut, überschwemmung, wirbelsturm, taifun, terremoto, volcán, deslizamiento, incendio, inundación, tifón, tremblement de terre, volcan, glissement de terrain, incendie, inondation, tornade, typhon & en, de, es, it & augmented CrisisLex data \\
flut, hochwasser, überschwemmung, inundation, flood, disaster, verstroming, hoogwater, vloed, inondation, crue, marée haute & de, en, nl, fr & 2021 Germany flood \\
incendio, forest fire, fuego forestal, bosque quemado & es, en & 2023 Chile forest fires \\
\bottomrule
\end{tabularx}
\label{tab:keywords}
\end{table}
\centering
\end{center}

Additionally, we used a fuzzy matching keyword-filtering method based on the Levenshtein string edit distance. It can be described using the recursive Formula~\ref{equation:levenshtein} where $x$ and $y$ represent two input strings and $i,j\in\mathbb{N}$ substring indices. The Levenshtein distance represents the minimum total cost required to transform one string into another string by applying a series of insertions, deletes and renames \cite{Levenshtein.1965}.

\begin{equation}
\begin{gathered}
    \text{ed}(x[1..0], y[1..0]) = 0 \\
    \text{ed}(x[1..i], y[1..0]) = i \\
    \text{ed}(x[1..0], y[1..j]) = j \\
    \text{ed}(x[1..i], y[1..j]) = \min \begin{cases}
        \text{ed}(x[1..i-1], y[1..j]) + 1 \\
        \text{ed}(x[1..i], y[1..j-1]) + 1 \\
        \text{ed}(x[1..i-1], y[1..j-1]) +\mathbbm{1}_{x[i] \neq y[j]} \\
    \end{cases}
\end{gathered}
\label{equation:levenshtein}
\end{equation}

The fuzzy matching filtering strategy was considered to account for minor spelling mistakes. A Tweet was identified as disaster-related whenever a keyword produced a fuzzy match with Levenshtein distance $\leq 2$.

\subsection{Generic fine-tuning}
Using the augmented generic data set from CrisisLex, a pre-trained \gls{RoBERTa} model was fine-tuned to provide a way to identify disaster-related Tweets without any manual labelling. It is referred to as the \gls{GFT} model going further. The underlying transformer architecture allows the model to consider words within their context using self-attention \cite{Vaswani.2017} and provides superior performance to other natural language processing approaches \cite{Devlin.2019}. For the implementation of this study, the multilingual Twitter-XLM-RoBERTa-base model by Barbieri et al. \cite{Barbieri.2022} was utilised as it is pre-trained on approximately $198$~million Tweets and has been shown to outperform the similar XLM-RoBERTa model trained using the more general CommonCrawl corpus \cite{Conneau.2020} for the multilingual classification of Tweets. The pre-trained model was fine-tuned for \numprint{11000}~iterations with batch size $8$ and an early-stopping strategy based on the validation loss.

\subsection{Active learning}
\label{sec:active_learning}
The \gls{AL} approach was undertaken using the unlabelled majority of the use case data. \gls{AL} addresses the lack of labelled training data for those scenarios and allows for the fine-tuning of the model's understanding of disaster-relatedness such that it aligns with the understanding of the user.

To test the effectiveness of \gls{AL} for the classification task in question, two configurations were considered: (1) taking the untouched Twitter-XLM-RoBERTa as a base model and (2) taking the version that was fine-tuned with generic data as described in the previous section as a base model. These two distinct setups were chosen to evaluate if fine-tuning with generic data helps to speed up the \gls{AL} process later on. For the main \gls{AL} loop, \numprint{10983} Tweets from the Ahr Valley flood and a uniformly selected sample of \numprint{20000} Tweets from the Chile forest fires were taken as input. The Chile data set was downsampled quite heavily to achieve bearable runtimes for each \gls{AL} iteration. Given this setup, \gls{AL} was conducted using a \gls{GCS} query strategy. It is based on a greedy computation of a subset of the unlabelled data points such that the geometric loss between the subset and the remaining data points is minimised \cite{Sener.2018}. 

\begin{figure}[ht]
\centering
    \includegraphics[width=.9\textwidth]{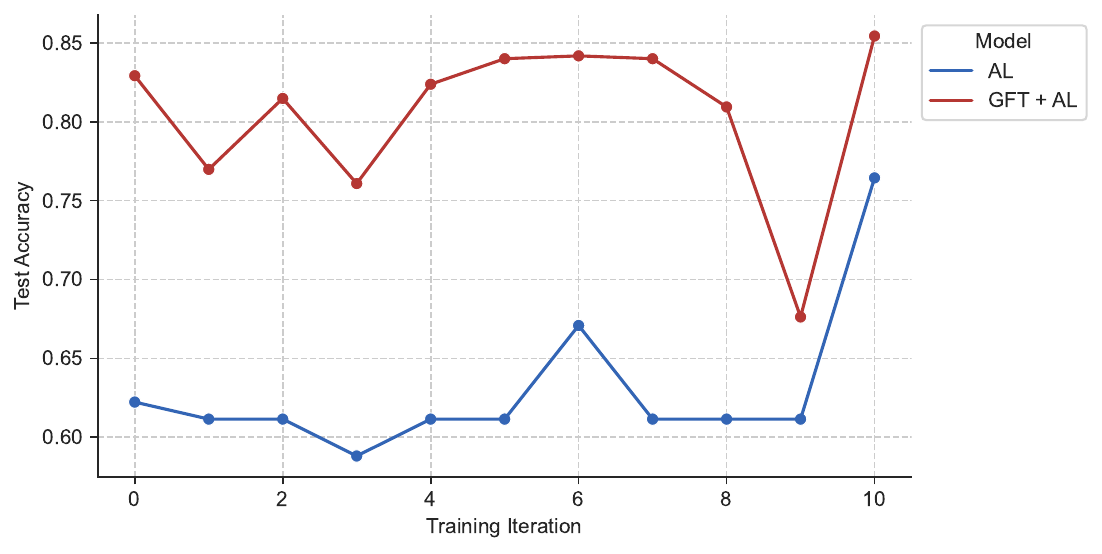}
    \caption{Test accuracy of the models trained (1) only with \gls{AL} and (2) with \acrfull{GFT} + \gls{AL} throughout the learning process}
    \label{fig:active_learning_accuracy}
\end{figure}

For both \gls{AL} setups, 10 rounds of querying data, labelling and updating were performed. During each iteration, 20 unlabelled data points were returned to the human annotators. Additionally, 20 data points were labelled to train a first instance of the model preceding the \gls{AL} process. Two persons participated in each part of the iterative labelling process to ensure the quality of the newly labelled training data. Fig.~\ref{fig:active_learning_accuracy} depicts the classification accuracy after each \gls{AL} iteration on the joint use case test data set from Germany and Chile.

\section{Results}
\label{sec:results}
Using standard evaluation metrics (accuracy, precision, recall, F1 score) \cite{Powers.2011}, we compared the outputs of our models. Table~\ref{tab:overview_results} shows the results for all our use cases: the Chile forest fires, the Western Germany flood and the generic data set derived from CrisisLex.




\begin{center}
\begin{table}[ht!]
\caption{Evaluation metrics for the data sets. The filtering strategies are abbreviated with \acrshort{KWF} for \acrlong{KWF}, \acrshort{GFT} for \acrlong{GFT}, and \acrshort{AL} for \acrlong{AL}. In the value pairs (e.g. 0.61 / 0.92), the left-hand value always stands for the "unrelated" class ($0$) and the right-hand value for "related" ($1$)}
\centering
\begin{tabular}{llllll}
\toprule
          & \textbf{\acrshort{KWF}} & \textbf{Fuzzy \acrshort{KWF}} & \textbf{\acrshort{GFT}} & \textbf{\acrshort{AL}} & \textbf{\acrshort{GFT} + \acrshort{AL}} \\
\midrule
\multicolumn{6}{c}{\textbf{\emph{(a) Evaluation metrics for CrisisLex}}} \\
\midrule
Precision & 0.61 / 0.92 & 0.64 / 0.85 & \textbf{0.96} / 0.92 & 0.53 / \textbf{0.95} & 0.94 / 0.94 \\
Recall   & 0.95 / 0.48 & 0.88 / 0.59 & 0.90 / \textbf{0.97} & \textbf{0.98} / 0.27 & 0.93 / 0.95 \\
F1 score  & 0.74 / 0.63 & 0.74 / 0.69 & \textbf{0.93} / \textbf{0.95} & 0.69 / 0.41 & \textbf{0.93} / 0.94 \\
Accuracy        & 0.70        & 0.72        & \textbf{0.94}        & 0.59 & \textbf{0.94} \\
\midrule
\multicolumn{6}{c}{\textbf{\emph{(b) Evaluation metrics for 2021~Germany flood}}} \\
\midrule
Precision & 0.96 / \textbf{1.00} & 0.96 / 0.86 & \textbf{0.98} / 0.77 & 0.94 / 0.82 & \textbf{0.98} / 0.87 \\
Recall   & \textbf{1.00} / 0.71 & 0.98 / 0.75 & 0.96 / \textbf{0.83} & 0.98 / 0.58 & 0.98 / \textbf{0.83} \\
F1 score   & \textbf{0.98} / 0.83 & 0.97 / 0.80 & 0.97 / 0.80 & 0.96 / 0.68 & \textbf{0.98} / \textbf{0.85} \\
Accuracy        & \textbf{0.96}        & 0.95        & 0.95        & 0.93 & \textbf{0.96} \\
\midrule
\multicolumn{6}{c}{\textbf{\emph{(c) Evaluation metrics for 2023~Chile forest fires}}} \\
\midrule
Precision  & 0.65 / 0.79 & 0.65 / 0.79 & 0.63 / 0.69 & 0.62 / 0.77 & \textbf{0.74} / \textbf{0.86} \\
Recall     & 0.82 / 0.61 & 0.82 / 0.61 & 0.67 / 0.65 & 0.82 / 0.55 & \textbf{0.87} / \textbf{0.73} \\
F1 score   & 0.73 / 0.69 & 0.73 / 0.69 & 0.65 / 0.67 & 0.71 / 0.64 & \textbf{0.80} / \textbf{0.79} \\
Accuracy        & 0.71        & 0.71        & 0.66        & 0.68 & \textbf{0.80} \\
\bottomrule
\end{tabular}
\label{tab:overview_results}
\end{table}

\centering
\end{center}

\section{Discussion}
\label{sec:discussion}
In this section, we first discuss the results of Sect.~\ref{sec:results} and then continue with a critical evaluation of the methodology, its limitations and an outlook on further research.

\subsection{Discussion of Results}
\label{subsec:discussion-of-results}
Keyword filtering performed relatively well for its simplicity but fell short compared to the other transformer-based approaches. For the generic test data set based on CrisisLex, the overall accuracy reached a value of 0.70. When looking at more detailed metrics, however, the method achieved very low recall values for class~1 ("related") and low precision for class~0 ("unrelated"). For the 2021~Germany flood data set, the approach worked best, even producing results that were comparable to the other filtering techniques. The results were also competitive for the 2023~Chile forest fire data set. Fuzzy keyword filtering only marginally improved the results of keyword filtering on the CrisisLex-based data set and barely resulted in a significant change in results in the other two data sets. For the 2023~Chile forest fire data, the results of hard and fuzzy keyword filtering were the same. For the purposes of this study, fuzzy keyword filtering thus did not yield any significant advantage.

Fine-tuning with generic data led to two-fold results. The classification performance for the respective CrisisLex-based test data set was high, yielding competitive values for all evaluation metrics. For the 2021~Germany flood test data, the \gls{GFT} approach produced slightly higher recall when compared to keyword filtering but lower precision for class~1. For the 2023~Chile forest fire data set, the performance was slightly worse throughout the board. Here, the model was even outperformed by the keyword-filtering approach. This might be traced back to the fact that out of 20~disasters covered by the generic training data set, only two were wildfires.


The approach based solely on \gls{AL} yielded unreliable results across the use cases. While its recall was very high for the CrisisLex data set for class~0 (0.98), it had a very low value for class~1 (0.27). The opposite was the case for the precision. It was noticeable that the scores were generally lower for class~1. The evaluation metrics were better for the Germany use case than for the Chile data set. Nevertheless, the \gls{AL} model achieved the lowest accuracy scores for all data sets. However, these results are limited by the number of \gls{AL} iterations and the relatively low number of data points labelled during each. This choice, though, was made on purpose in this study, as a lengthy and computationally expensive labelling process would defeat the purpose of \gls{AL} for the rapid identification of disaster-related Tweets.  

Our approach combining \gls{GFT} and \gls{AL} scored the highest evaluation metrics. This was particularly the case for the Chile use case, where the model consistently outperformed all other approaches. For the Germany use case, the performance was considerably higher, although class~1 was around 10~percentage points worse than class~0. For the CrisisLex data set, the model performed slightly worse than the plain \gls{GFT} model, with the exception of precision for class~1 and recall for class~0. Still, the lowest overall metric was a recall of 0.73 for class~1 in Chile, which was 8~percentage points higher than the lowest value in the plain \gls{GFT} model. Given these results, \gls{AL} significantly improved upon the \gls{GFT} approach, especially for the Chile use case, and yielded a model that could be broadly applied to different disaster scenarios.

\subsection{Discussion of Methodology}
\label{subsec:discussion-of-methodology}
Although the filtering methods compared in this study are popular and established techniques for social media and text analysis, the choice of methods is not exhaustive. Numerous other methods have been utilised for this task including \glspl{CNN} \cite{Sufi.2022}, \glspl{LSTM} \cite{Huang.2022} and \glspl{GNN} \cite{Papadimos.2023}. Naturally, keyword filtering and transformer-based classification approaches are vastly different and have been developed in different time frames. However, keyword filtering still constitutes a simple and quick method for filtering or querying textual data. Compared to modern approaches, a keyword filter requires much less computational power and can easily be applied to large amounts of data residing in e.g. a database without further technical considerations. 
However, it comes with the fundamental problem of keyword selection. In addition, morphological restrictions (e.g. due to inflection), the multilingualism of a data set and the resulting polysemantics pose a particular challenge. Both problems are addressed by our \gls{AL} and \gls{GFT} approaches which do not require the identification of keywords and can handle multiple languages. Nevertheless, the results of this study show that keyword filtering can yield high performance for specific data when it has been investigated carefully and keywords are selected accordingly.

The \gls{GFT} approach described in this study requires zero additional labelling effort and still leverages the advantages of modern \glspl{LLM}. However, it must be noted that such a generic data set might not be available for every use case. Although CrisisLex covers a fairly wide range of scenarios, some events such as avalanches might not be reflected in the generic training data, limiting the model's suitability for such scenarios. In such a case, manual labelling would once again be necessary, defeating the purpose of fine-tuning with generic data. On the other hand, \gls{GFT} holds a large potential for transfer learning which is leveraged in this study by subsequent \gls{AL} and making the classification model multilingual. While the \gls{GFT} model trained for this study can theoretically also be applied to texts written in Non-Indo-European languages, the performance and transferability of a fine-tuned Twitter-XLM-RoBERTa model on such languages are subject to further studies. A study by Zheng et al. \cite{Zheng.2022} has already shown large differences in accuracy of approximately 15~percentage points between Indo-European and Non-Indo-European languages.

\gls{AL} also comes with its unique challenges. There are numerous query strategies and none of them has been proven to be generally superior to the other. On a use case basis, different query strategies can yield varying results. The query strategy must therefore be carefully chosen by the user, though every strategy is better than random sampling \cite{Ein-Dor.2020}. During the implementation of this study, the \gls{GCS} strategy provided superior results compared to other strategies such as \gls{LC} and also came with better runtimes. It was therefore chosen for the final evaluation.
The \gls{AL} process conducted for this study was furthermore quite inconsistent as depicted by Fig.~\ref{fig:active_learning_accuracy}. The pure \gls{AL} model dropped in accuracy for the first few rounds and only yielded competitive performance after the tenth round. The case was similar for the combined \gls{AL} + \gls{GFT} approach with performance losses for the first few rounds and slight increases afterwards. It also experienced a sharp drop in test accuracy after the ninth round of \gls{AL} and a subsequent sharp increase. The F1 scores experienced similar inconsistencies and even dropped to 0.00 for the first few rounds of the pure \gls{AL} approach.

Lastly, the field of language processing is changing rapidly and novel methods for few-shot classification \cite{Schroder.2023} and zero-shot classification with the help of generative \glspl{LLM} \cite{Wang.2023} have rapidly gained popularity. Although these methods have not yet generally outperformed more traditional classification approaches and \gls{AL}, they are subject to future investigations. In the context of \gls{AL}, generative \glspl{LLM} might also be useful for automating the annotation of training examples.

\section{Conclusion}
In this paper, we compared an \acrfull{AL}-based approach to identify disaster-related Tweets with keyword filtering and generic fine-tuning approach using \gls{RoBERTa}. For evaluation, we compared three data sets: a generic collection of Tweets from CrisisLex, Tweets from the 2021~West Germany Flood, and Tweets from the 2023~Chile forest fires.

We found that simple keyword filtering produces good results in many cases, but is fundamentally inferior to RoBERTa-based methods. A fuzzy approach did not improve our keyword filtering-based results.
We observed clear differences between our use cases. The classification of the Tweets from Germany generally worked better than for the Spanish-language data from Chile. Our plain \gls{AL} model performed the worst out of all approaches, though, for some parts of the data, the differences to some of the other techniques were small, especially for the Chile data set.
The approach combining \gls{AL} with an already fine-tuned RoBERTa model achieved the best performance across most evaluation metrics. In particular, for the Chile use case, it significantly outperformed the generically fine-tuned model, which scored even worse than simple keyword filtering. The approach was also superior to all other methods for the Germany use case and produced competitive results when applied to generic test data based on CrisisLex.

Consequently, we can verify that combining \gls{AL} with a generic fine-tuning approach is a well-suited strategy for the identification of disaster-related Tweets. The approach required very little labelling of data and outperformed all other methods for the data tested in our study.

\section*{Acknowledgements}
This preprint has not undergone peer review or any post-submission improvements or corrections. The Version of record of this contribution is published in the Springer series \href{https://www.springer.com/series/15179}{Lecture Notes in Networks and Systems}, and is available online at \url{https://doi.org/10.1007/978-3-031-66428-1_8}.

This research has furthermore received funding from the European Commission - European Union under HORIZON EUROPE (HORIZON Research and Innovation Actions) under grant agreement 101093003 (HORIZON-CL4-2022-DATA-01-01). Views and opinions expressed are however those of the author(s) only and do not necessarily reflect those of the European Union - European Commission. Neither the European Commission nor the European Union can be held responsible for them. This work has also received funding from the Austrian Federal Ministry for Climate Action, Environment, Energy, Mobility, Innovation and Technology (BMK) project GeoSHARING (Grant Number 878652).

\bibliographystyle{plain}
\bibliography{bibliography.bib}

\end{document}